\documentclass{article}

\PassOptionsToPackage{numbers, compress}{natbib}

\usepackage[final]{neurips_2020}

\usepackage[utf8]{inputenc} 
\usepackage[T1]{fontenc}    
\usepackage{hyperref}       
\usepackage{url}            
\usepackage{booktabs}       
\usepackage{amsfonts}       
\usepackage{nicefrac}       
\usepackage{microtype}      
\usepackage{graphicx}
\usepackage{enumitem}

\title{Unsupervised in-distribution anomaly detection of new physics through conditional density estimation}

\author{%
  George Stein\thanks{Corresponding author}\\
 Berkeley Center for Cosmological Physics\\
 University of California, Berkeley\\
 Lawrence Berkeley National Laboratory\\
 Berkeley, CA, USA\\
  \texttt{gstein@berkeley.edu} \\
  \And
   Uro\v{s} Seljak\\
 Berkeley Center for Cosmological Physics\\
 University of California, Berkeley\\
 Lawrence Berkeley National Laboratory\\
 Berkeley, CA, USA\\
 \texttt{useljak@berkeley.edu}
   \And
   Biwei Dai \\
 Berkeley Center for Cosmological Physics\\
 University of California, Berkeley\\
 Berkeley, CA, USA\\
  \texttt{biwei@berkeley.edu}
}
\begin{document}

\maketitle

\begin{abstract}
Anomaly detection is a key application of machine learning, but is generally focused on the detection of outlying samples in the low probability density regions of data. Here we instead present and motivate a method for unsupervised in-distribution anomaly detection using a conditional density estimator, designed to find unique, yet completely unknown, sets of samples residing in high probability density regions. We apply this method towards the detection of new physics in simulated Large Hadron Collider (LHC) particle collisions as part of the 2020 LHC Olympics blind challenge, and show how we detected a new particle appearing in only 0.08\% of 1 million collision events. The results we present are our original blind submission to the 2020 LHC Olympics, where it achieved the state-of-the-art performance. 
\end{abstract}

\section{Introduction}

The detection of anomalous data is a fundamental problem in machine learning with applications to many scientific disciplines. Most commonly, anomaly detection is synonymous with outlier detection and the identification of Out-of-Distribution (OoD) examples \citep{OOD_1, OOD_2, OOD_3, OOD_4}. Applications typically involve the search for samples in low probability density areas of the data near the tails of various data distributions, in order to remove or flag them for further inspection. Density estimation techniques are often 
used for this application, but have been shown to be 
problematic in high dimensions, where out-of-distribution data can achieve higher density than in-distribution data \cite{Nalisnick2019}. 
Recently a number of alternatives have 
been proposed that alleviate this issue \cite{LikelihoodRatioAI,PAE}. 

Alternatively, a problem may instead motivate the search for {\textit{in-distribution anomalies}} - loosely defined as a small set of samples that reside in areas of the data with high probability density, but have unique yet unknown properties when compared to their surroundings. If the given data is expected to be smoothly distributed, in-distribution anomalies may present themselves as local over-densities in a region of the parameter space. In this regime it is no longer desirable to only determine if the probability density of a sample is high or low, we instead wish to compare the density of a sample to that of its neighbours along some conditional dimension of interest. Such unsupervised in-distribution anomaly searches are ideal for scientific applications with large amounts of data where the observable space, parameter space, or model space may be too large to perform directed searches for some set of already-known signatures, and blind searches are required instead. 

One exciting application is to look for evidence of new particles produced in high-energy collisions at the LHC. To date, directed searches targeting specific signals have yet to find evidence for new particles \citep{model_search}, leading to a broader desire to ensure that the LHC search program is available to capture an unknown rare signal in a broader and less supervised search. For this purpose, the {\textit{2020 LHC Olympics}}\footnote{\url{https://lhco2020.github.io/homepage/}} were held. A simulated dataset was released in advance, containing 1 million events meant to be representative of actual LHC observations. Participants competed to detect an anomalous event in the `black box' data set (only the first of three black boxes was the focus of the winter Olympics), without any knowledge of what the signal may look like.

This work approaches the LHC signal detection challenge as an example of in-distribution anomaly detection. The strategy we pursue is to look for excess density in a narrow region of a parameter of interest, such as the invariant mass of each event. In this paper we perform conditional density estimation with Gaussianizing Iterative Slicing (GIS) \cite{sig}, and construct an in-distribution anomaly score to detect the signal in a completely blind manner. Our technique achieved state-of-the-art results, correctly identifying the anomalous events and the physics of the signal. The results presented in this submission are unchanged from our blind submission to the LHC Olympics in January 2020. Parallel and independent to our development and application of our conditional density estimation method, a similar one was applied in \cite{anode}.

\section{Dataset and data preprocessing}
\label{sec:data}
We used the publicly available datasets from the \textit{LHC Olympics 2020 Anomaly Detection Challenge}. The R\&D dataset \cite{lhc_randd} was used for constructing and testing the method, while the first of the `black boxes' \cite{lhc_bb1} was the basis of our submission to the winter Olympics challenge. The black box contains 1 million events simulated with Pythia 8 \cite{pythia1, pythia2}, meant to be realistic representations of the LHC collisions. The data for each event consists of the four momenta for each of the up to 700 particles measured, determined through a detector simulation using Delphes 3.4.1 \cite{delphes}. The particles are recorded in detector coordinates ($p_T, \eta, \phi$). As the particles are the result of hadronic decays we expect them to be spatially clustered in a number of jets. By focusing on the jet summary statistics rather than the particle data from an event we are able to vastly reduce the dimensionality of the data space. 
We used the python interface of FastJet \cite{fastjet} - pyjet \cite{pyjet} - to perform jet clustering, setting R=1.0 as the jet radius and keeping all jets with $|\eta| < 2.5$. Each jet $J$ is described by a mass $m_J$, a linear momentum $p=(p_T, \eta, \phi)$, and n-subjettiness ratios $\tau^J_{n n-1}$ \cite{subjettiness1,subjettiness2}, which describe the structure and number of sub-jets within each jet. A pair of jets has an invariant mass $M_{JJ}$. To construct images of the jets we binned each particles transverse momentum $p_T$ in $(\eta,\phi)$ and oriented using the moment of inertia. For the final black box 1 run we limited events to 2250 GeV < $M_{JJ}$ < 4750 GeV in order to remove all samples in low density regions, resulting in 744,217 events remaining after all data cuts. 

\section{Method}

Our in-distribution anomaly detection method relies on a framework for conditional density estimation. Current state-of-the-art density estimation methods are those of flow-based models, popularized by \cite{realnvp} and comprehensively reviewed in \cite{normalizing_flows}. A conditional normalizing flow (NF) aims to model the conditional distribution $p(x|x_c)$ of input data $x$ with conditional parameter $x_c$  by introducing a sequence of $N$ differentiable and invertible transformations $f = f_1 \circ f_2 \circ \dots \circ f_N $ to a random variable $z$ with a simple probability density function $\pi (z)$, generally a unit Gaussian. Through the change of variables formula the probability density of the data can be evaluated as the product of the density of the transformed sample and the associated change in volume introduced by the sequence of transformations: 
\begin{equation}
    \label{eq:density}
    p(x|x_c) = \pi(f_{x_c}(x)) \left| \mathrm{det} \left(\frac{\partial f_{x_c}(x)}{\partial x} \right) \right| =  \pi(f_{x_c}(x)) \prod_{i=1}^{i=N} \left| \mathrm{det} \left(\frac{\partial f_{x_c,i}(x)}{\partial x} \right) \right| .
\end{equation}
While various NF implementations make different choices for the form of the transformations $f_i$ and their inverse $f_i^{-1}$, they are generally chosen such that the determinant of the Jacobian, $\mathrm{det} (\partial f_{x_c,i}(x)/\partial x)$, is easy to compute. Mainstream NF methods follow the deep learning paradigm: parametrize the transformations using neural networks, train by maximizing the likelihood, and optimize the large number of parameters in each layer through back-propagation.

In this work we use an alternative approach to the current deep learning methodology, a new type of normalizing flow - Gaussianizing Iterative Slicing (GIS) \cite{sig}. GIS works by iteratively matching the 1D marginalized distribution of the data to a Gaussian. 
At iteration $i$, the transformation of data $X_i$, $f_{x_c,i}$, can be written as
$X_{i+1} = X_i - W_iW_i^TX_i + W_i \mathbf{\Psi}_{x_c,i}(W_i^T X_i)$,
where $W_i$ is the weight matrix that satisfies $W_i^TW_i=I$, and $\mathbf{\Psi}_{x_c,i}$ is the 1D marginal Gaussianization of each dimension of $W_i^T X_i$.
To improve the efficiency, the directions of the 1D slices $W_i$ are chosen to maximize the PDF difference between the data and Gaussian using the Wasserstein distance at each iteration. The conditional dependence on $x_c$ is modelled by binning the data in $x_c$ and estimating a 1D mapping $\mathbf{\Psi}_i$ for each $x_c$ bin, then interpolating ($W_i$ is the same for different $x_c$ bins). 
GIS can perform an efficient parametrization and calculation of the transformations in Equation~\ref{eq:density}, with little hyperparameter tuning. We expect that standard conditional normalizing flow methods would also work well for this task, but did not perform any comparisons. 

\begin{figure}[h]
  \centering
  \includegraphics[width=0.85\textwidth]{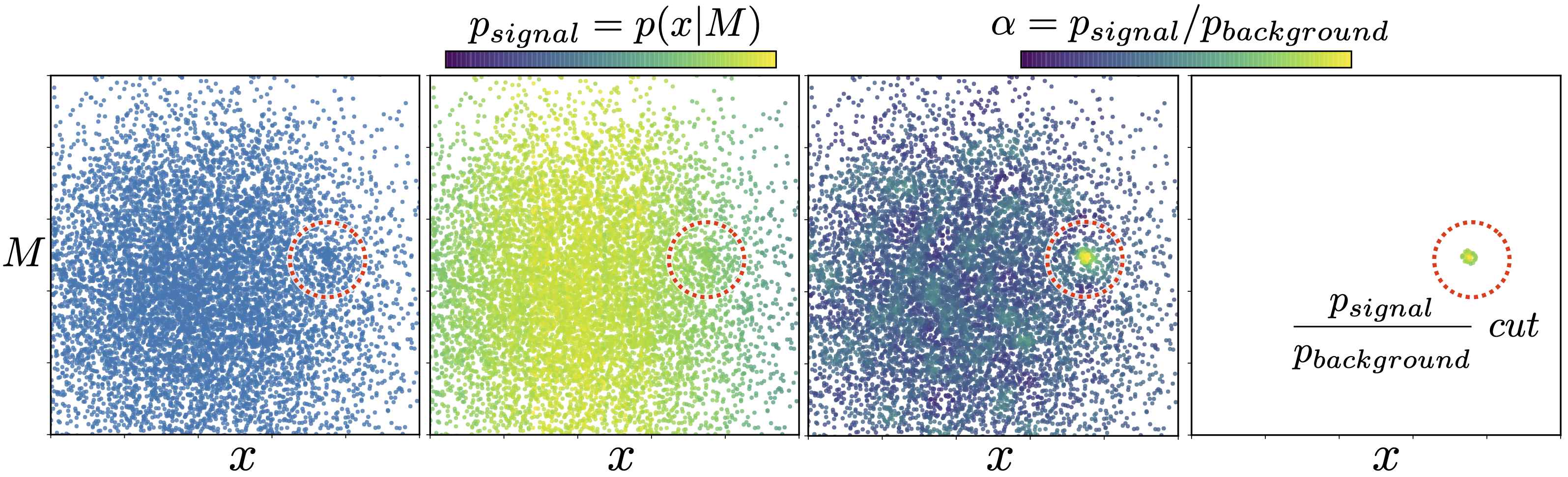}
  
  \vspace{-7pt}
  \caption{In-distribution anomaly detection through conditional density estimation. Consider samples of a 1D feature $x$ and a conditional parameter of interest $M$ (left panel), drawn from a smooth Gaussian `background' with a small number of anomalous `signal' events added (inside red circle for clarity). The conditional density values at each data point do not allow the anomaly to be distinguished from the background (center left panel), as they only identify the outliers in low density regions. However, the local over-density anomaly ratio $\mathrm{\alpha}$ peaks at the anomalous data points (center right panel), and implementing a minimum cut on the anomaly ratio reveals the anomalous events (right panel).}
  \label{fig:method}
\end{figure}

With the GIS NF trained to calculate the conditional density, our in-distribution anomaly detection method, illustrated in Figure~\ref{fig:method}, works as following: 
\vspace{-5pt}
\begin{enumerate}[leftmargin=*]
    \itemsep-0em 
    \item Calculate the conditional density at each data point $p(x|M_{JJ})$, denoting this $\mathrm{p_{signal}}$, using the jet masses and n-subjettiness ratios as the data $x$ and the invariant mass of a pair of jets $M_{JJ}$ as the conditional parameter. For our application the invariant mass was chosen as we are searching for new particles, and hence we expect them to have a specific mass. For other scientific applications the conditional parameter will vary, and needs to be chosen by hand using domain-specific knowledge, or all parameters can be iterated over.  
    \item Calculate the density at neighbouring regions along the conditional dimension, $p(x | M_{JJ} \pm \Delta)$, and interpolate to get a density estimate in the absence of any anomaly. This is denoted $\mathrm{p_{background}}$. Explore various values of $\Delta$. 
    \item The local over-density ratio $\mathrm{\alpha = p_{signal}/p_{background}}$ will be $\approx1$ in the presence of a smooth background with no anomaly. 
A sign of an anomalous event is $\alpha>1$. 
    Individual events can also be selected based on the desired $\alpha$ characteristic.
\end{enumerate}
\vspace{-10pt}

\vspace{-12pt}
\section{Results}

If there is an anomalous particle decay in the data its jet decay products would likely be located in a narrow range of masses, corresponding to the mass of the particle itself. For this reason we chose the invariant mass $M_{JJ}$ of two jets as the conditional parameter to conduct the anomaly search along. We iterated on selections of jets $i$ and $k$, and selections of n-subjettiness ratios, and found the most significant anomaly when investigating the lead two jets and the first n-subjettiness ratio, so we used \{$M_{JJ}$, $m_{J_1}$, $m_{J_1}-m_{J_2}$, $\tau_{21}^{J_1}$,  $\tau_{21}^{J_2}$\} as the 5 parameters describing each event. We also experimented with training a convolutional autoencoder on the jet images, reasoning that rare events (anomalies) would have a higher reconstruction error and different latent space variables than more common ones, as seen in \cite{ae_error}. While we found this to be true on the R\&D dataset, the autoencoder-based variables introduced more noise in the  density estimation than the physics-based parameters, so they were not used here. 

Simple investigations of the dataset showed that it was smoothly distributed, and no anomalies were apparent by eye. We trained the conditional GIS on all events, and evaluated the anomaly score $\alpha$ for each datapoint. On the R\&D set we found that point estimates of the conditional densities resulted in a larger noise level than convolving the conditional density with a Gaussian PDF of width $\sigma=\Delta$ (1-PDF convolution for the background), discretely sampled at 10 points, so used the Gaussian-convolved probability estimates. $\mathrm{\sigma=250\ GeV}$ provided the most strongly peaked signal. 

\begin{figure}[h]
  \centering
  \includegraphics[width=0.54\textwidth]{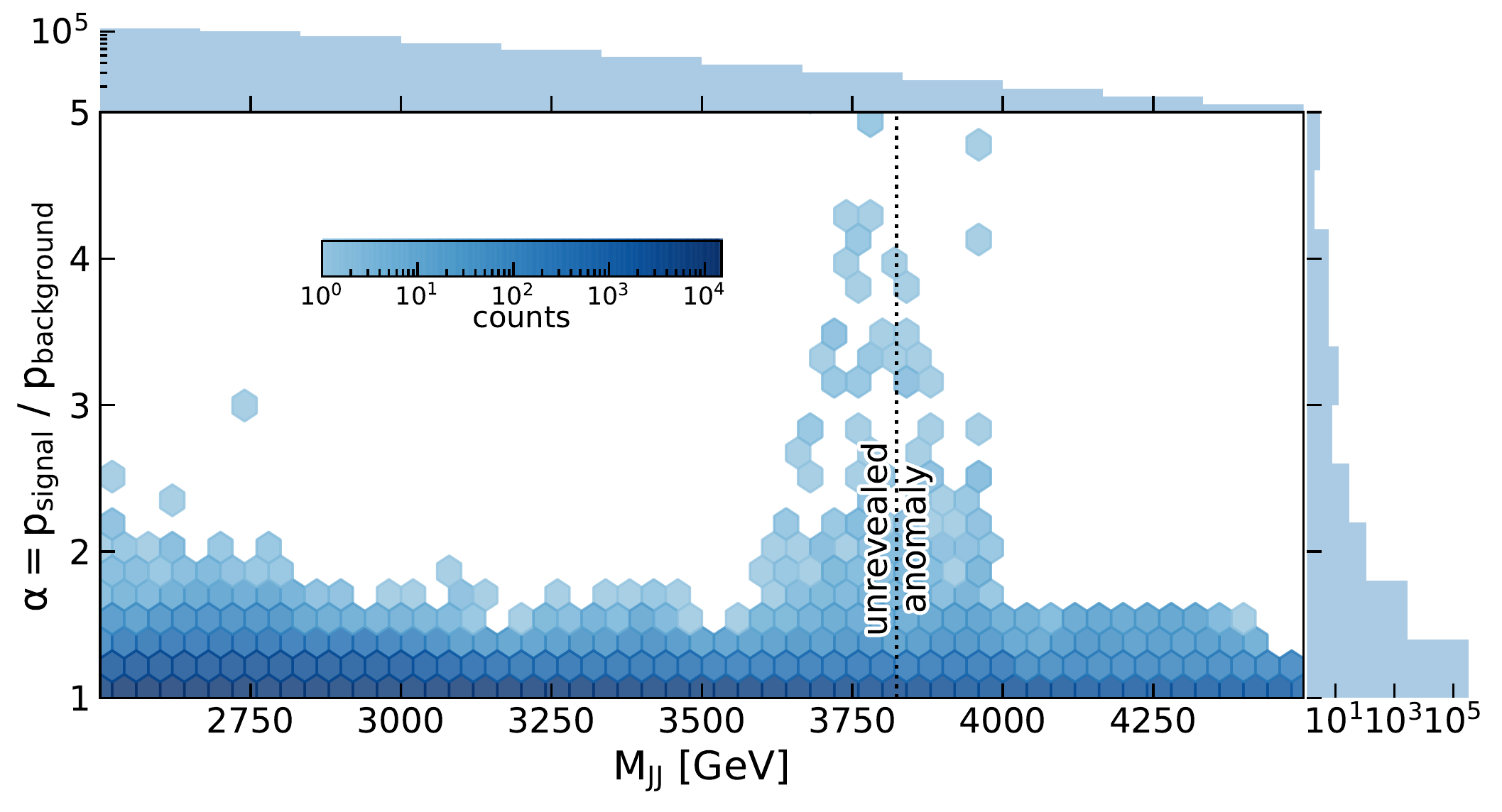}
\includegraphics[width=0.45\textwidth]{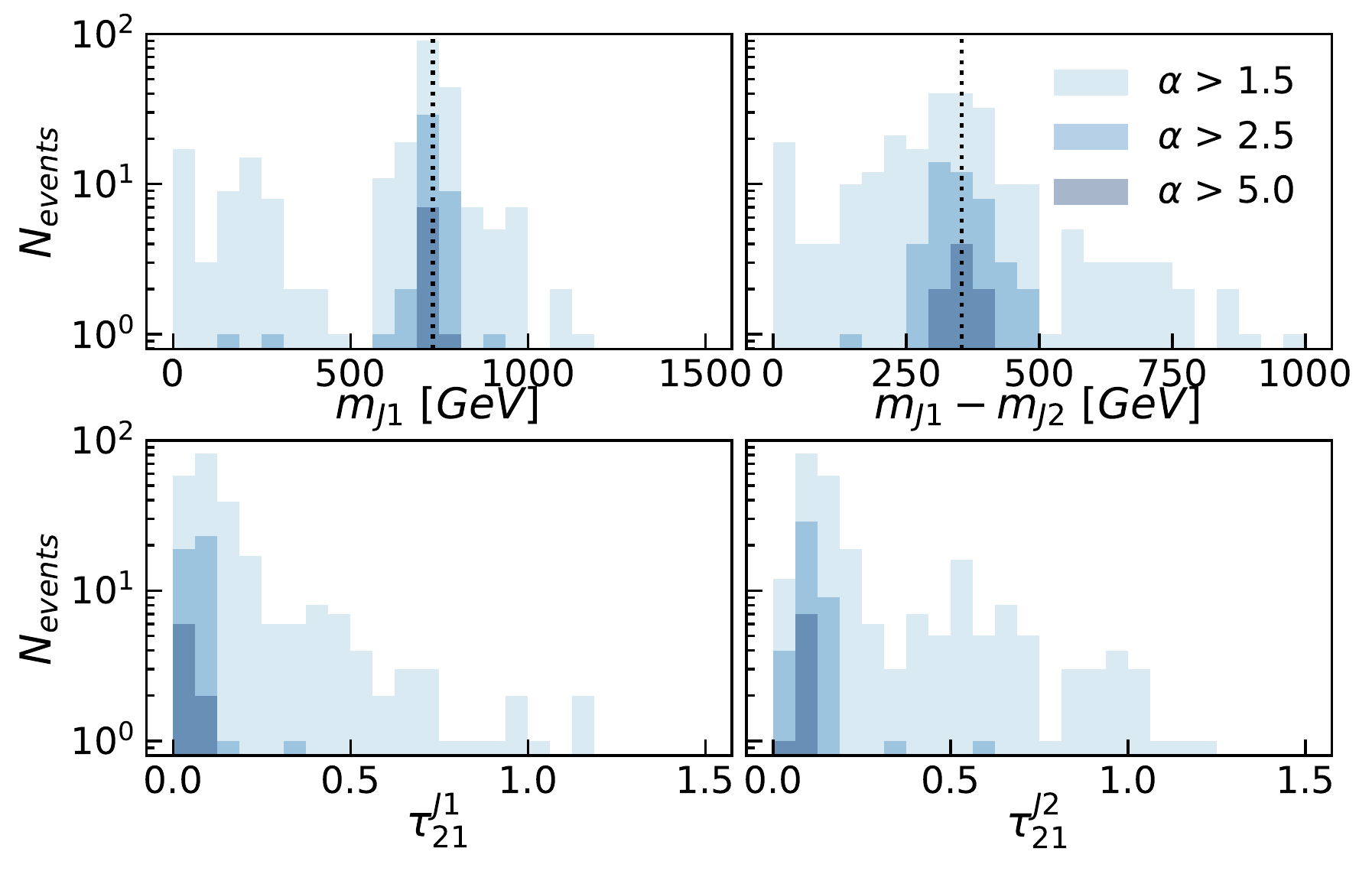}

  \vspace{-11pt}
  \caption{{\bf Left}: The anomaly score for each event as a function of the invariant mass of the leading two jets. A number of anomalous events are clearly seen near $\mathrm{M_{JJ}\approx 3750 GeV}$. {\bf Right:} parameter distributions of the events that remain after imposing cuts on the anomaly score $\alpha$. Vertical dashed lines are the true anomalous events that were unveiled after the close of the competition.}
  \label{fig:histograms}
\end{figure}

As seen in the left panel of Figure~\ref{fig:histograms}, the anomaly score strongly peaks around $\mathrm{M_{JJ}\approx 3750\ GeV}$. If these events are truly from a particle decay we expect that their resulting jet statistics will be clustered around some mean value, unlike if it is simply a result of noise in the model or background. To investigate the apparent anomaly we remove data outside of $\mathrm{3600\ GeV < M_{JJ} < 3900\ GeV}$, and look at the events that remain after a series of cuts on the anomaly score $\alpha$. We show the parameter distributions of the events that remain after imposing $\alpha > [1.5, 2.5, 5.0]$ cuts in the right four panels, and find that the most anomalous events are centered in $M_{J1}$ and $M_{J1}-M_{J2}$, and have small values of n-subjettiness $\tau_{21}$. This strongly indicates that we found a unique over-density of events that do not have similar counterparts at neighbouring $M_{JJ}$ values - i.e. an anomaly. While here we show hand-selected values of the anomaly score cuts chosen to result in narrowing parameter histograms for illustrative purposes, they can also be chosen a priori as multiples of the standard deviation of $\alpha$ across the dataset, or the standard deviation in each mass bin, to similar effect.  

\vspace{-11pt}
\begin{figure}[h]
  \centering
  \includegraphics[width=0.56\textwidth]{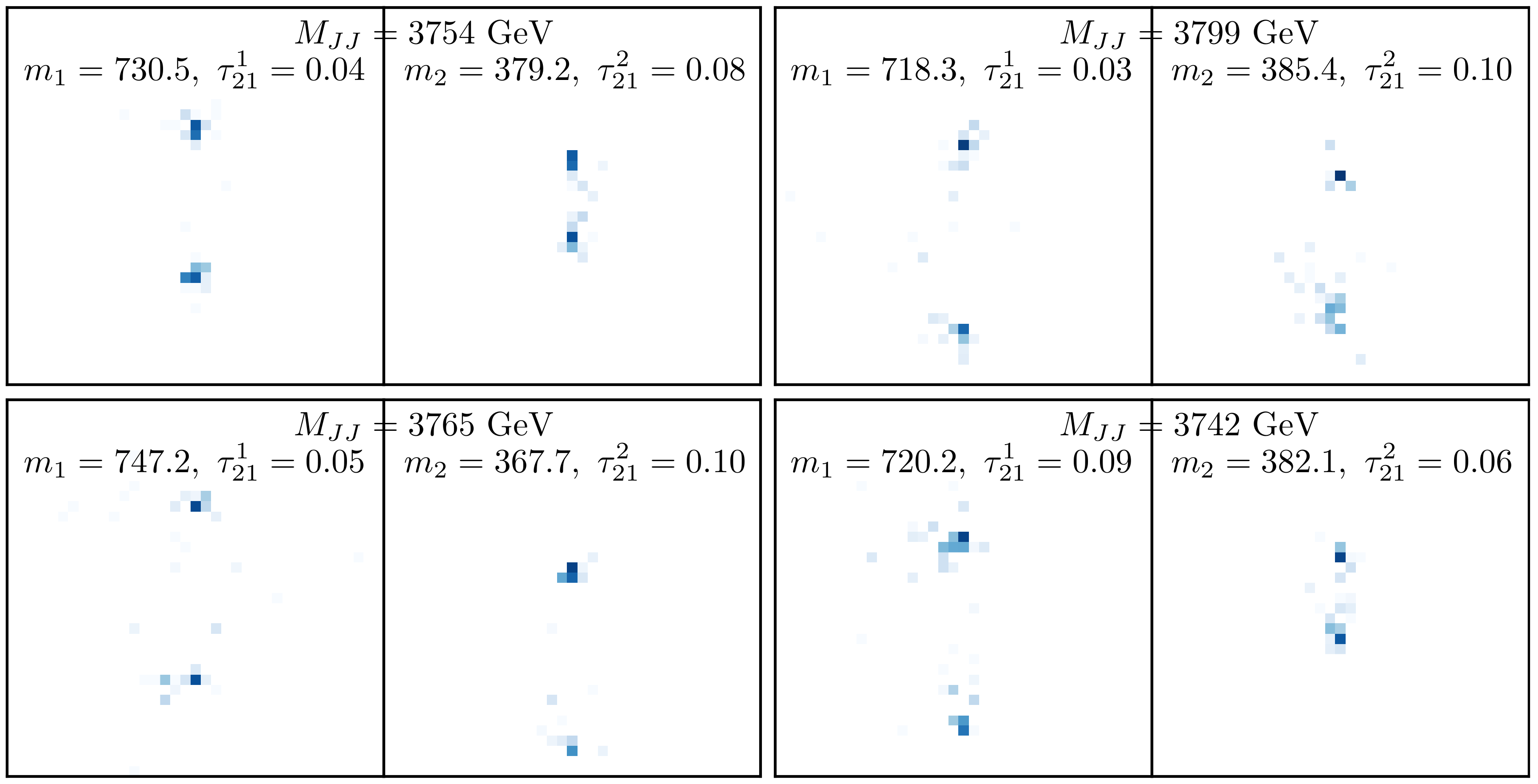}
    \includegraphics[width=0.43\textwidth]{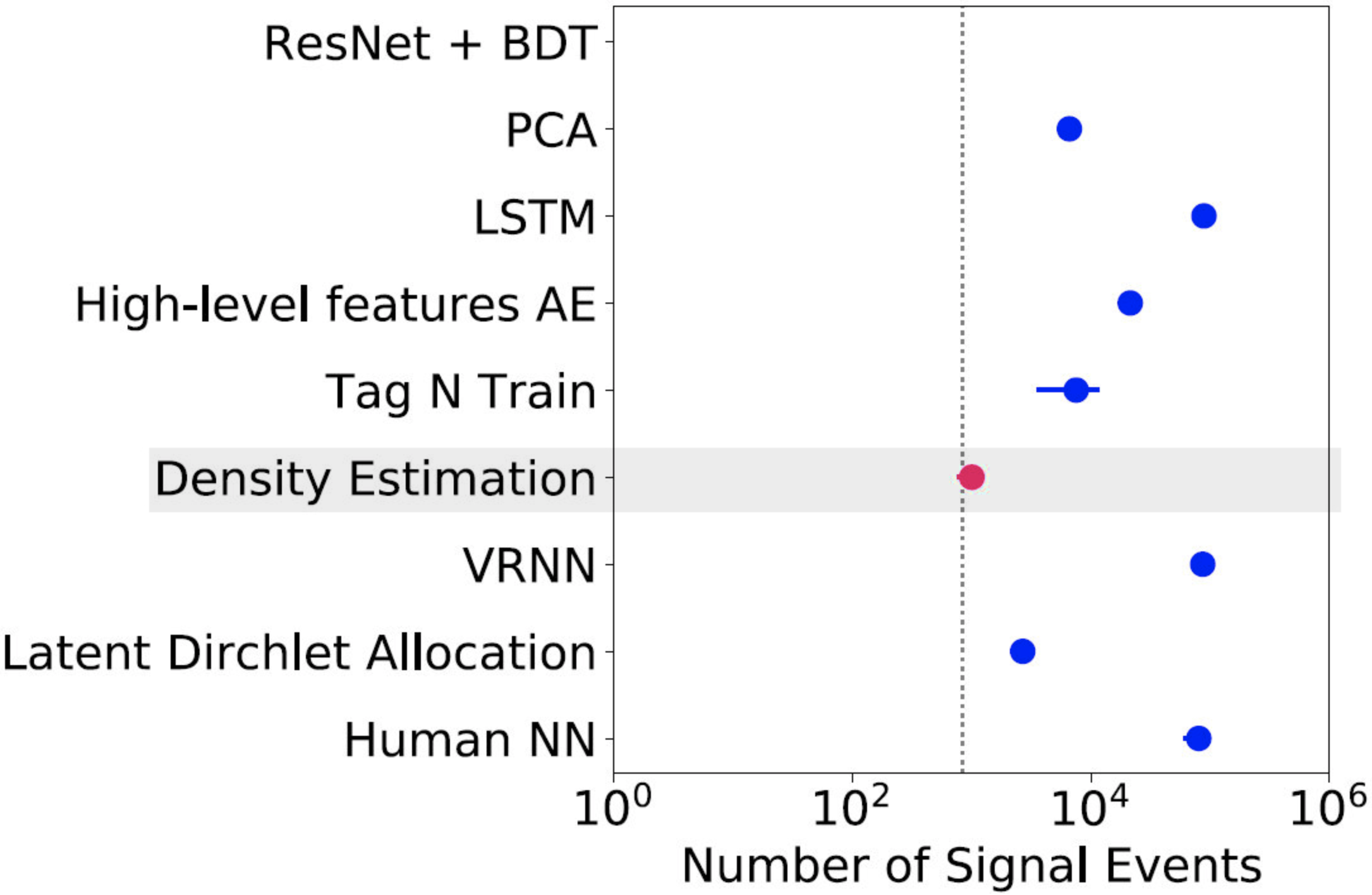}
  \vspace{-17pt}
  \caption{{\bf Left:} The four most anomalous events in the black box. Each pair of images visualizes the particles belonging to the lead two jets. {\bf Right:} Number of signal events as predicted by the different submissions to LHC Olympics 2020. The vertical dashed line is the correct answer, and our density estimation submission is shown by the red dot. See \cite{lhc_results} for more details.}
  \label{fig:images}
\end{figure}

We visualized the events ranked by decreasing anomaly score in Figure~\ref{fig:images}, and found that each of the leading two jets for events with a high anomaly score additionally have very similar visual appearances. Using the events that remain after an $\alpha > 2.0$ cut we summarized the anomalous events as follows: a $\mathrm{3772.9\pm 8.3\ GeV}$ particle decays into 2 particles, one with $\mathrm{M_{1}=727.8 \pm 3.8\ GeV}$, and the other with $\mathrm{M_2 = 374.8 \pm 3.5\ GeV}$. Each of these decayed into  two-pronged jets. This is in great agreement with the truth revealed after the close of the competition: a 3823 GeV particle decays into two particles of 732 GeV and 378 GeV. Based on the corresponding analysis of the R\&D data we estimated that there were a total of $1000\pm 200$ of these events included in the black box of a million total events. Our estimate was within one sigma of the true answer of 834 events (Figure \ref{fig:images}).
\vspace{-5pt}
\section{Conclusion}
We trained a conditional density estimator on the first of the black boxes released for the 2020 LHC winter Olympics. Using the conditional density estimation we constructed a local over-density method for anomaly detection. 
Application of the method to the blind
challenge of LHC Olympics 2020 
revealed a 
very good agreement between the 
predictions and true values after they were revealed at the close of the competition,, with a state-of-the-art performance achieved in comparison to other methods. 
The success of the conditional density estimation for the in-distribution anomaly detection in realistic LHC data suggests it may be useful more broadly. It is particularly 
suitable for analyzing  low or high dimensional data for which the background is expected to vary smoothly, while 
the anomaly is localized 
in a specific way, such as localized in 
one or several variables. A good 
example is astronomical data, where 
the anomaly may be localized spatially or may have specific signatures in the time series data. 


   
\section*{Broader Impact}

Scientists may benefit from this research in a variety of domains, and there may be industrial applications. We see no potential for unethical applications of the method or future societal consequences. There are no identifiable disadvantages of this work, and system failures and biases in the data are not applicable to this work.

\bibliographystyle{abbrvnat}
\bibliography{references}

\end{document}